\documentclass[conference]{IEEEtran}
\ifCLASSOPTIONcompsoc
  \usepackage[nocompress]{cite}
\else
  \usepackage{cite}
\fi
\usepackage{amsfonts}
\usepackage{wrapfig}
\usepackage[font={small}]{caption}
\ifCLASSINFOpdf
 \usepackage[pdftex]{graphicx}
 \DeclareGraphicsExtensions{.pdf,.jpeg,.png}
\else
   \DeclareGraphicsExtensions{.eps}
\fi
\usepackage{pgfplots}
\usepackage{algorithmic}
\usepackage{algorithm}
\usepackage[fleqn]{amsmath}
\usepackage{amsmath}
\usepackage{tabularx}
\usepackage{graphicx}
\usepackage{multicol, blindtext}
\usepackage{breqn}
\usepackage{multirow}
\usepackage{subfig}
\usepackage{breqn}
\usepackage{color}
\usepackage{graphics}
\usepackage{colortbl}
\usepackage{tabulary}
\usepackage{etoolbox}
\usepackage{colortbl}
\usepackage{bigstrut}
\usepackage{multirow}
\usepackage{array}
\usepackage{tabularx}
\usepackage{booktabs}
\usepackage{hhline}
\usepackage{dsfont}

\usepackage[a4paper]{geometry}
\newgeometry{right=0.62in,left=0.62in,top=0.75in,bottom=1in}

\title{Dynamic Distribution of Edge Intelligence at the Node Level for Internet of Things.}

\vspace{0mm}

\author{
    \IEEEauthorblockN{Hawzhin Mohammed\IEEEauthorrefmark{1}, Tolulope A. Odetola\IEEEauthorrefmark{1}, Nan Guo\IEEEauthorrefmark{7}, Syed Rafay Hasan\IEEEauthorrefmark{1}}

    \vspace{0mm}
    \IEEEauthorblockA{\IEEEauthorrefmark{1}Department of Electrical and Computer Engineering, Tennessee Tech University, Cookeville, TN, 38505}
    \IEEEauthorblockA{\IEEEauthorrefmark{7}Center for Manufacturing Research, Tennessee Tech University, Cookeville, TN, 38505}
\vspace{0mm}
}

\begin{document}
\maketitle
\vspace{0mm}
\begin{abstract}
In this paper, dynamic deployment of Convolutional Neural Network (CNN) architecture is proposed utilizing only IoT-level devices.
By partitioning and pipelining the CNN, it horizontally distributes the computation load among resource-constrained devices (called as horizontal collaboration), which in turn increases the throughput.
Through partitioning, we can decrease the computation and energy consumption on individual IoT devices and increase the throughput without sacrificing accuracy.
Also, by processing the data at the generation point, data privacy can be achieved.
The results show that throughput can be increased by 1.55x to 1.75x for sharing the CNN into two and three resource-constrained devices, respectively.

\end{abstract}


\vspace{0mm}
\section{Introduction}
\label{Introduction}
\vspace{0mm}

\footnote[1]{A version of this work will be published in MWSCAS 2021} Internet of Things (IoT) devices are widespread in the smart city, smart grid, advanced manufacturing, health monitoring, and other modern applications.
The IoT device combines different sensors, objects, and smart devices that can communicate with each other without human interference \cite{ambrosin2016feasibility}.
The IoT devices perform autonomously in conjunction with other devices.
Cisco Inc. forecasts that 500 billion devices will be connected to the internet by 2030 \cite{Cisco2016}.
Each IoT device collects data through its sensors, communicates with its surroundings, and over the internet with other devices.
A network of these connected IoT devices forms the IoT network.
Data generated by these smart connected IoT devices used in the analysis to assist make more informed decisions and actions \cite{conti2018internet}.

As the number of IoT devices increases, the number of data collected by them also increases.
This data needs to be sent to the data center for analysis and decision-making.
Sending an enormous amount of data over the internet cause delay, bandwidth occupation, power consumption, and raise privacy issues.
To solve these problems, data need to be processed at the location of the generation \cite{zhou2018robust}.
Edge computing is a model which brings data processing and data storage closer to the location where it is generated, to improve response times, bandwidth utilization, energy consumption, and to preserve privacy \cite{ding2017trunk}.

\begin{figure}[h]
\vspace{0mm}
\setlength{\abovecaptionskip}{2mm}   
\setlength{\belowcaptionskip}{0mm}   
\centering
\includegraphics[width=0.41\textwidth]{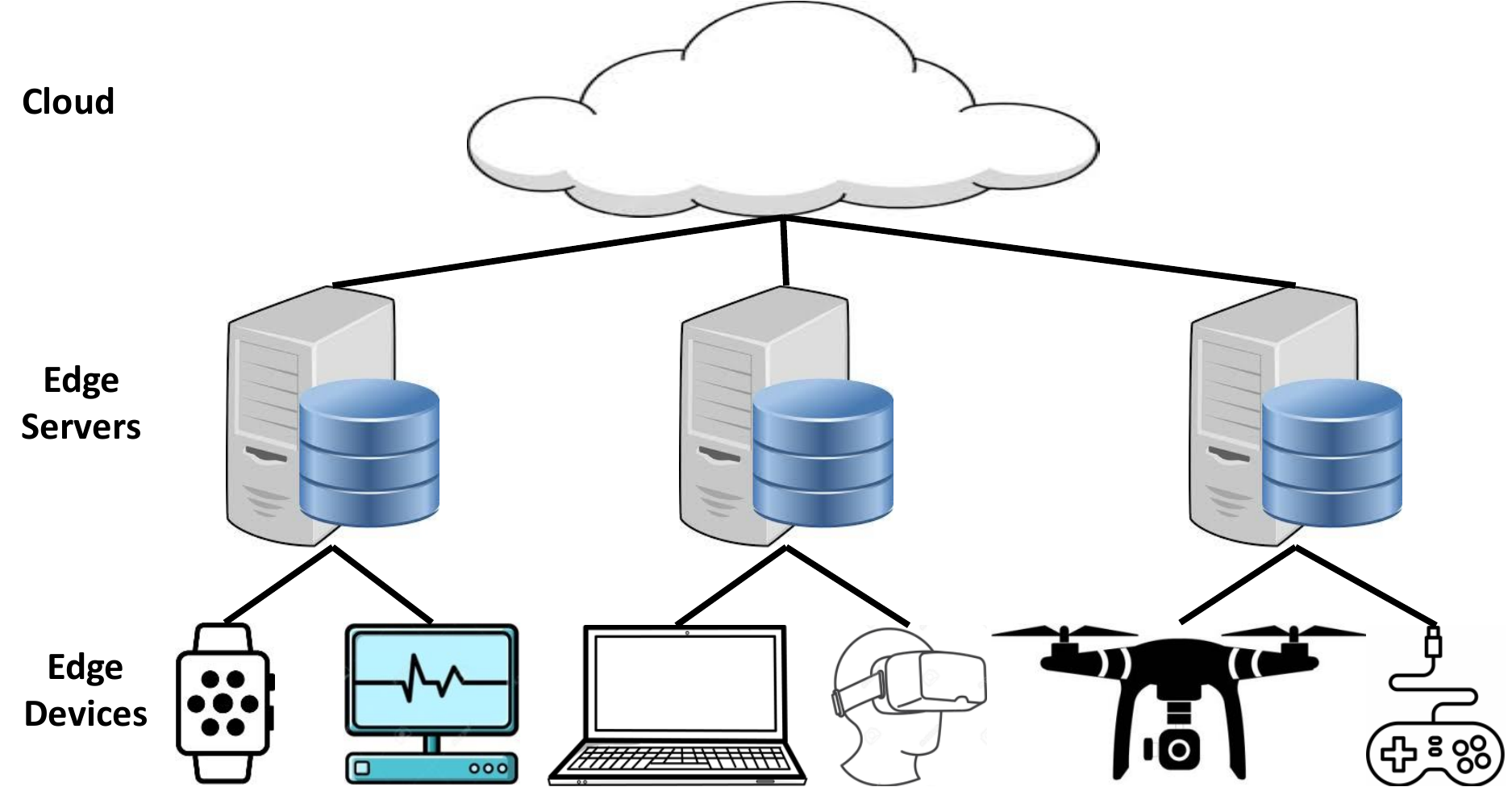}
\caption{Hierarchy of Edge Device Communication}
\vspace{0mm}
\label{cloud}
\end{figure}

Convolutional Neural Network (CNN) in recent times has become a state-of-the-art machine learning technique and has gained adoption in fields like computer vision, natural language processing, object detection among others.
The huge computation and memory requirement of CNN makes their deployment on resource-constrained edge devices (e.g. IoT devices) a challenging task \cite{li2018edge}.
Many research works have proposed different approaches to ensure edge devices accommodate computational and memory requirements of CNN.
Traditional techniques include cloud computing which involves the generation of input images from edge devices which are transmitted to the cloud for inference and sent back to edge devices \cite{li2018edge} as seen in Fig. \ref{cloud}.
This approach has raised privacy concerns with the drawback of high latency \cite{li2018edge, zhao2018deepthings}. 

Another approach involves model compression using techniques such as pruning, parameter quantization, Huffman encoding, and hash function. This approach achieves a reduction in computing workloads but leads to a reduction in classification accuracy \cite{mao2017modnn}.
The resolution of latency issues has inspired edge computing technologies leading to Edge-based Intelligence (EBI) which alleviates and migrates computation tasks from the cloud to the network edges resulting in a reduction in network communication overhead, lower latency, and a more energy-efficient CNN inference compared to cloud computing\cite{li2018edge}.

EBI adds some amount of latency between edge devices and edge servers as shown in Fig. \ref{cloud}.
EBI is also dependent on the available bandwidth between the edge device and the edge server, as bandwidth decreases, inference's latency increases especially in volatile bandwidth applications \cite{li2018edge}.
Moreover, applications such as smart city security, traffic management, or disaster relief in remote and humanly inaccessible areas, require to take decisions more time which is very challenging for resource-constrained devices. 
Hence, to improve the performance of EBI of CNNs in terms of latency, the inference of CNNs can be done between edge devices excluding edge servers thereby improving latency.
Also, the network can be made dynamically and computation required for intelligence can be distributed among multiple devices (horizontal collaboration).

In this paper, dynamic partitioning methodology (DPM) is explored for the optimal distribution of the layers of CNN among edge devices to achieve faster and accurate inference.

\begin{figure}[t]
\vspace{0mm}
\setlength{\abovecaptionskip}{1mm} 
\setlength{\belowcaptionskip}{0mm} 
\centering
\includegraphics[width=0.43\textwidth]{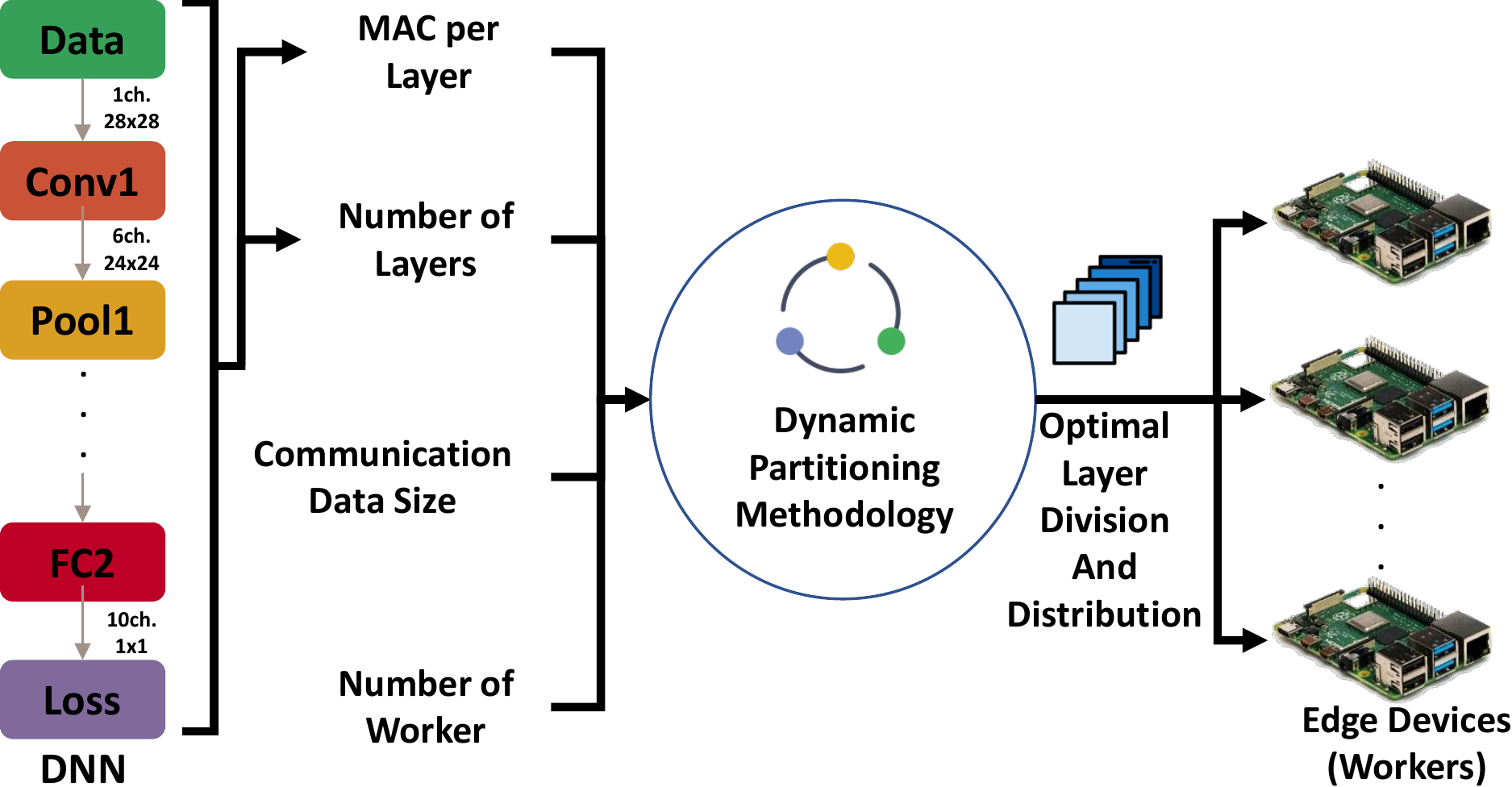}
\caption{Proposed Methodology}
\vspace{0mm}
\label{fig:Methodology}
\end{figure}

The main contribution of the paper is as follows:
\begin{itemize}
    \item Dynamic distribution of intelligence among resource-constrained devices.
    \item Insight on how to optimally divide the CNN among different IoT devices or workers to increase the throughput.
\end{itemize}

The remainder of this paper is organized as follows:
The proposed methodology is presented in Section \ref{Proposed Methodology}.
Section \ref{How to Optimally Divide a CNN} is about how to optimally divide a CNN network.
Experimental setup is presented in Section \ref{Experimental Setup}.
Section \ref{Results and Discussions} presents the results.
Comparison with state-of-the-art is presented in Section \ref{Comparison With State-of-the-Art}.
Conclusions are drawn in Section \ref{Conclusion}.

\begin{algorithm}
\caption{Dynamic partitioning methodology - DPM}
\begin{algorithmic} [1]
\STATE \textbf{Input to DPM:}
\STATE Number of MAC operations per layer
\STATE Number of layers in the CNN
\STATE Communication channel bandwidth
\STATE Number of workers
\STATE \textbf{Operation:}
\STATE Calculate the total MAC operations in the CNN
\STATE Divide the result by the number of the worker devices
\STATE Divide the CNN such that each part has approximately the same number of MAC operations
\STATE Calculate the data size of each part
\IF{Output data size $>$ Communication channel bandwidth}
\STATE Move the division location to decrease the output size
\ENDIF
\STATE Distribute the parts among the worker devices
\STATE \textbf{Output:} Optimal layer division and distribution
\end{algorithmic}
\vspace{0mm}
\label{alg:MYALG}
\end{algorithm}
\vspace{0mm}
\section{Proposed Methodology}
\label{Proposed Methodology}
\vspace{0mm}

In this section, the proposed methodology is explained.
First, some terminology that has been used in this paper needs to be defined.
An edge device or the IoT device is the device that collects data at the edge of the network.
The worker is an edge device that shares its resources with other edge devices (workers).
The requester is the edge device that uses other workers' resources to speed up the inference process.
Also, the requester is responsible for creating the network and provide the data (image) for inference.
The inputs to DPM are the number of workers, size of data that need to be sent over the communication channel, number of layers, and number of MAC operations per layer (line 2 to 5 in Algorithm \ref{alg:MYALG}).
Algorithm \ref{alg:MYALG} and Fig. \ref{fig:Methodology} summarize the proposed methodology.

Initially, the algorithm calculates the total number of MAC operations by multiplying the number of layers by the number of MAC operations per layer (line 7).
Then, it calculates the number of MAC operations per worker by dividing the total number of MAC operations by the number of workers (line 8).
After that, the algorithm divides the CNN such that each worker gets an equal number of MAC operations (line 9).
Then, it calculates the data size of each part and compares it with communication channel bandwidth, if the communication channel cannot handle the calculated data size then the algorithm moves the division location to the next layer (lines 10, 11, and 12).
Finally, the algorithm distributes parts among workers (line 14).

In this way, the DPM guarantees no channel data overflows no worker computation overflows, and maximizes the throughput.
This algorithm also makes sure that each worker can process the data that it receives.

\begin{table}[t]
\vspace{0mm}
  \centering
  \caption{Number of MAC operations per layer and output size of each layers in LeNet CNN.}
    \begin{tabular}{cccc}
    \toprule
    \hspace{0mm} Layer \# \hspace{0mm} & \hspace{5mm} Layer type \hspace{5mm} & \hspace{0mm} MAC operation \hspace{0mm} & \hspace{0mm} Output size \hspace{0mm} \\
    \midrule
I & Convolution & 86400 & 3456 \\
II & Max Pooling & 3460 & 864 \\
III & Convolution & 153600 & 1024 \\
IV & Max Pooling & 1020 & 256 \\
V & Convolution & 30720 & 120 \\
VI & Fully Connected & 10080 & 84 \\
VII & Fully Connected & 840 & 10 \\
    \midrule
    \end{tabular}%
    \vspace{0mm}
  \label{tab:lenet_MAC}%
\end{table}%
\begin{table}[t]
  \centering
  \caption{Inference time needed for one worker, two worker and three worker to process 100 images.}
    \begin{tabular}{cccc}
    \toprule
    \hspace{0mm} Case \hspace{0mm} & \hspace{0mm} \# worker \hspace{0mm} & \hspace{0mm} Time/image (ms) \hspace{0mm} & \hspace{-1mm} Communication time (ms) \hspace{-1mm} \\
    \midrule
I & One & 5.40 & 0 \\
II & Two & 3.48 & 2.13 \\
III & Three & 3.09 & 2.13 and 1.65 \\
    \midrule
    \end{tabular}%
    \vspace{0mm}
  \label{tab:time}%
\end{table}%

\vspace{0mm}
\section{How to Optimally Divide a CNN}
\label{How to Optimally Divide a CNN}
\vspace{0mm}

One important metric of deployment CNN on a hardware unit is the number of multiply-accumulate (MAC) operations.
As the number of MAC operations increases the computation process increases and consequently, the time to finish inference of an image increases.
For example, LeNet has seven layers \cite{lecun1998gradient}, as shown in Fig \ref{fig:OneWorker}.
Three of these layers are convolution, two of them are max pooling, and the final two are fully connected.
Table \ref{tab:lenet_MAC} shows the number of MAC operations per layer (Table \ref{tab:time} shows the advantages of distributed edge intelligence, further discussed in Section \ref{Results and Discussions}).
The convolution layers have higher MAC operations compared to other types of layers \cite{hailesellasie2019mulnet}.
To optimally divide a CNN, the number of MAC operations has to be near equally divided among the number of participated devices, i.e., workers.

But the number of MAC operations for pipeline operations is just half of the story. The other half is the amount of the data (layer output) that needs to be sent over the communication channel.
When a CNN runs on a single device, the amount of data that is sent from one layer to the other does not make much difference i.e. minimal time is spent in transferring data between layers.
Whereas for pipeline operations where a communication channel involves between the workers, the layer's output size determines the latency due to the communication, as shown in Figs \ref{fig:TwoWorker} and \ref{fig:ThreeWorker}.
The size of the data that needs to be sent from one worker to another depends on the location where the partition of CNN is made.
For instance, in Fig. \ref{fig:OneWorker}, if the CNN has been divided such that conv1 and pool1 be on one worker and the rest be on another then, in this case, the first worker needs to send 864 of data to the other worker.
But, if the CNN has been divided such that conv1, pool1, conv2, and pool2 be on worker one and the rest of the CNN are deployed on other workers, then first worker requires to send 256.

To pipeline the CNN, two metrics must be taken into consideration, the number of MAC operations in each device and the size of the data that needs to be transmitted (as shown in Table \ref{tab:time}).
The algorithm is optimized to ensure that an equal number of operations occur at each worker with minimal data output or data communication size.
Choosing minimal communication data to be sent, increases the CNN's throughput.

\vspace{0mm}
\begin{equation*}
Pipeline \hspace{3pt} time \propto Number \hspace{3pt} of \hspace{3pt} MAC \hspace{3pt} operations
\end{equation*}
\vspace{0mm}
\begin{equation*}
Pipeline \hspace{3pt} latency \hspace{3pt} time/stage \propto Size \hspace{3pt} of \hspace{3pt} output
\end{equation*}
\vspace{0mm}

DPM searches for optimal combination among the number of workers, MAC operation, and data communication size, using Algorithm 1, to divide the CNN in the location that guarantees equal MAC distribution with minimal communication data.

\vspace{0mm}
\section{Experimental Setup}
\label{Experimental Setup}
\vspace{0mm}

In this section, experimental setup is explained that is used to conduct experiments to confirm the effectiveness of DPM.

\vspace{0mm}
\subsection{Device Type(s)}
\label{Device Type}
\vspace{0mm}

The testbed consists of Raspberry Pi 3 Model B, which is the third generation of Raspberry Pi.
Its specification is: 1) quad-core 1.2GHz Broadcom BCM2837 64bit CPU, 2) 1GB RAM, 3) BCM43438 wireless LAN and Bluetooth Low Energy (BLE) on board, and 4) 100 base Ethernet interface.
These devices are utilized because they are one of the most commonly used platforms in modern IoT-based network experiments \cite{hamdan2019iot, pavithra2015iot, vujovic2015raspberry}.

\begin{figure}[t]
\vspace{0mm}
\setlength{\abovecaptionskip}{0mm}   
\setlength{\belowcaptionskip}{0mm}   
\centering
\vspace{0mm}
\subfloat[Full LeNet run on one worker.
    \label{fig:OneWorker}]
    {\includegraphics[width=1.0\linewidth]{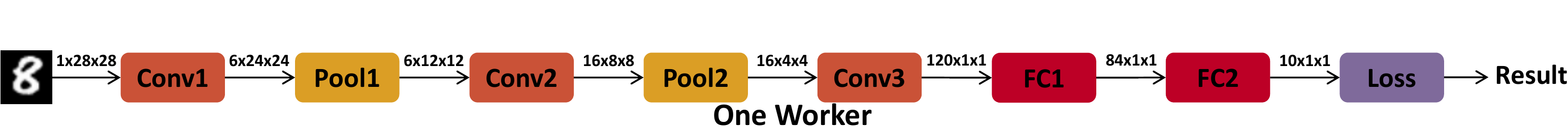}}
    \hfill
    \vspace{0mm}
    
\subfloat[Full LeNet run on two workers.
    \label{fig:TwoWorker}]
    {\includegraphics[width=1.0\linewidth]{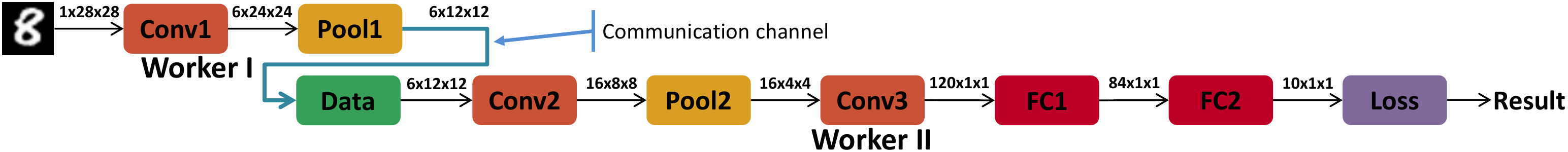}}
    \hfill
    \vspace{0mm}
    
\subfloat[Full LeNet run on three workers.
    \label{fig:ThreeWorker}]
    {\includegraphics[width=1.0\linewidth]{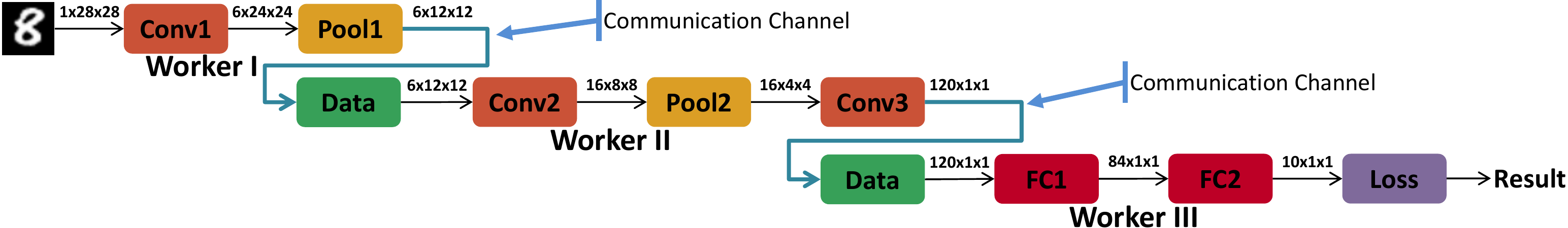}}
    \hfill
    \vspace{0mm}
    
\caption{Normal case and Attacks type on HAN network}
\vspace{0mm}
\end{figure}

\vspace{0mm}
\subsection{Experimental Scenarios}
\label{Experimental Scenarios}
\vspace{0mm}

There are three scenarios, namely, 1) full LeNet CNN on a single worker device, 2) LeNet CNN is divided between two different worker devices, and 3) LeNet CNN is divided among three different worker devices.
As shown in Fig. \ref{fig:OneWorker}, in the first scenario there is one case which is the full LeNet running on a single worker device.
Table \ref{tab:lenet} shows all scenarios and cases.

The second scenario has six cases.
The LeNet has been divided at different locations (layer) between the two worker devices.
As shown in Table \ref{tab:lenet}, in the first case the LeNet has been distributed between the two workers, where the first worker runs only the first convolutional layer (conv1) and the rest of the CNN runs on the second worker.
In the second case, the first worker runs the first two layers (conv1 and pool1), and the rest of the CNN is runs on the second worker, as shown in Fig. \ref{fig:TwoWorker}.
The partitioning of the CNN continue in this way and finally, in the last case, the first worker runs all the layer of LeNet CNN except for last fully connected layers is handled by the second worker.

In the third scenario, as shown in Table \ref{tab:lenet}, the first two layers (conv1 and pool1) are run on the first worker, conv2 and pool2 are run on the second and the rest of the layers are run on the third worker.
The second case is conv1 and pool1 is run on the first worker, conv2, pool2, and conv3 are run on the second and the rest of the layers are run on the third worker, as shown in Fig. \ref{fig:ThreeWorker}.

\begin{table}[t]
\vspace{0mm}
  \centering
  \caption{Scenarios and cases of CNN distribution among works.}
    \begin{tabular}{ccccc}
    \toprule
    \hspace{-1mm} Scenario \hspace{-1mm} & \hspace{-1mm} Case \hspace{-1mm} & \hspace{-1mm} Worker I \hspace{-1mm} & \hspace{-1mm} Worker II \hspace{-1mm} & \hspace{-1mm} Worker III \hspace{-1mm} \\
    \midrule
I & 1 & Full LeNet & -- & -- \\
    \midrule
\multirow{6}{*}{II} & 1 & conv1 & pool1 to ip2 & -- \\
                    & 2 & conv1 \& pool1 & conv2 to ip2 & -- \\
                    & 3 & conv1 to conv2 & pool2 to ip2 & -- \\
                    & 4 & conv1 to pool2 & conv3 to ip2 & -- \\
                    & 5 & conv1 to conv3 & ip1 \& ip2 & -- \\
                    & 6 & conv1 to ip1 & ip2 & -- \\
    \midrule
\multirow{2}{*}{III} & 1 & conv1 \& pool1 & conv2 \& pool2 & conv3 to ip2 \\
                     & 2 & conv1 \& pool1 & conv2 to conv3 & ip1 \& ip2 \\
    \midrule
    \end{tabular}%
    \vspace{0mm}
  \label{tab:lenet}%
\end{table}%
\begin{table}[t]
  \centering
  \caption{Throughput percentage of one, two, and three workers.}
    \begin{tabular}{cccc}
    \toprule
    \hspace{2mm} Case \hspace{2mm} & \hspace{2mm} \# worker \hspace{2mm} & \hspace{2mm} Throughput \hspace{2mm} & \hspace{2mm} Time (ms) \hspace{2mm} \\
    \midrule
I & One & 100\% & 540.103 \\
II & Two & 155\% & 347.780 \\
III & Three & 175\% & 308.457 \\
    \midrule
    \end{tabular}%
    \vspace{0mm}
  \label{tab:throughput}%
\end{table}%

\vspace{0mm}
\section{Results and Discussions}
\label{Results and Discussions}
\vspace{0mm}

In this section, two types of operations have been discussed, non-pipeline and pipeline operations.
In a non-pipeline (case I), there is one worker that contains all the LeNet CNN.
In pipeline (case II and case III), the LeNet CNN distributed between the workers to increase the throughput (horizontal collaboration).
As it can be seen from Table \ref{tab:throughput}, pipeline cases outperform non-pipeline case.
Also, when the number of workers increases, the throughput increases, and the time to inference decreases.
For the two workers in Table \ref{tab:time} and Table \ref{tab:throughput}, DPM chooses the best performance case among the six cases in Table \ref{tab:lenet} to compare with the best two cases of three workers.

From Table \ref{tab:throughput}, when the LeNet CNN run on one worker (non-pipeline) to classify 100 images, the device took more than 540 ms.
To classify the same number of images using pipeline operation, with two workers it took about 348 ms and for three workers the time decreased to about 308 ms.
Using a pipeline for inference can increase the throughput up to 155\% for two workers and up to 175\% for three workers.

For single image inference, as shown in Table \ref{tab:time}, in the non-pipeline case when there is only one worker, to classify one image it took 5.4 ms.
The communication time is zero as there is no other device involved in the classification.
In case II, the communication time between the two worker nodes to transmit 3456 (as the LeNet CNN divided at pool1) took 2.13 ms.
To classify an image, the two workers need an extra 2.13 ms added to the inference time compare to the non-pipeline.
For 100 images the communication time is 213 ms added to the 540 ms for the normal inference which becomes 753 ms, but because of the pipelining, the time for inference decreases to 348 ms.
These decrements in time are due to the pipelining operation as each worker does its data processing without waiting for the previous worker.
The decrements are even more substantial with case III when three workers cooperate to classify the images.
Even with the increase of the number of worker nodes which lead to increase in the number of communication link and consequently communication latency, the classification time decreases.

\begin{figure}[t]
\vspace{0mm}
\setlength{\abovecaptionskip}{0mm}
\setlength{\belowcaptionskip}{0mm}
\centering
\subfloat[Comparison between one worker node and two worker nodes for 100 to 1,000 images.
    \label{fig:2S1K}]
    {\includegraphics[width=0.49\linewidth]{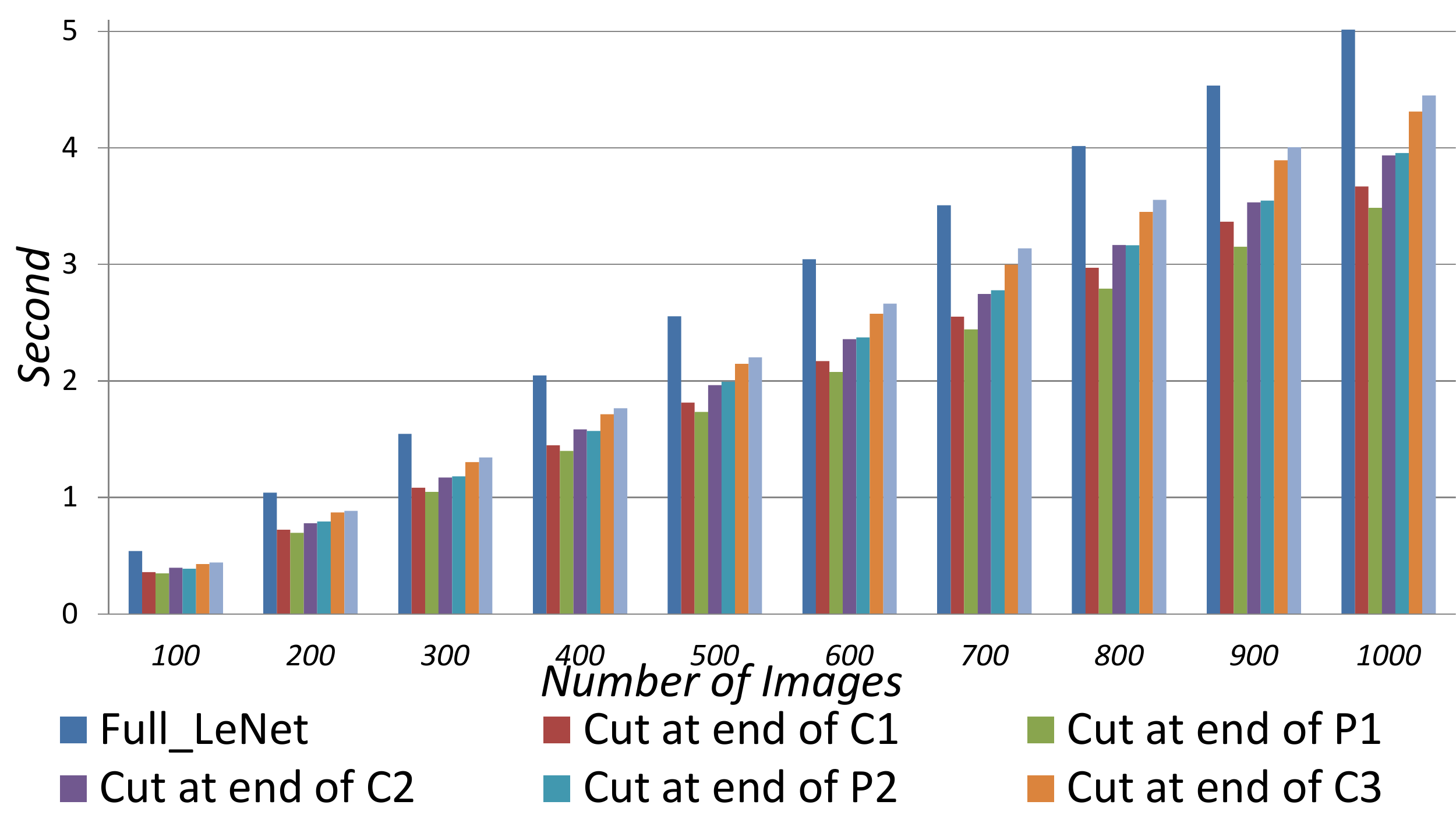}}
    \hfill
\subfloat[Comparison between one worker node and two worker nodes for 1,000 to 10,000 images.
    \label{fig:2S10K}]
    {\includegraphics[width=0.49\linewidth]{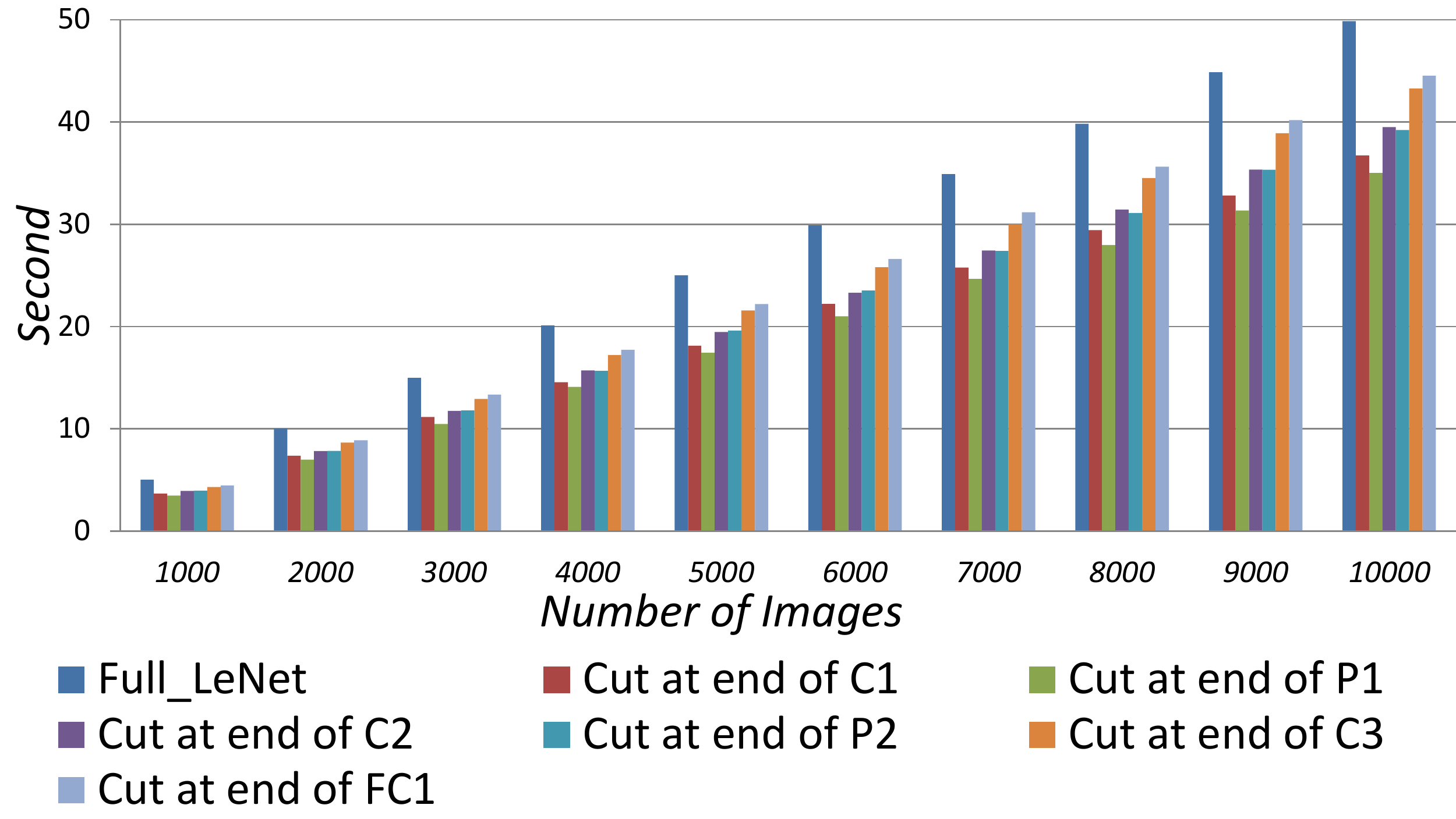}}
    \hfill
    \vspace{0mm}
\subfloat[Comparison among worker node, two worker nodes, and three worker nodes for 100 to 1,000 images.
    \label{fig:3S1K}]
    {\includegraphics[width=0.49\linewidth]{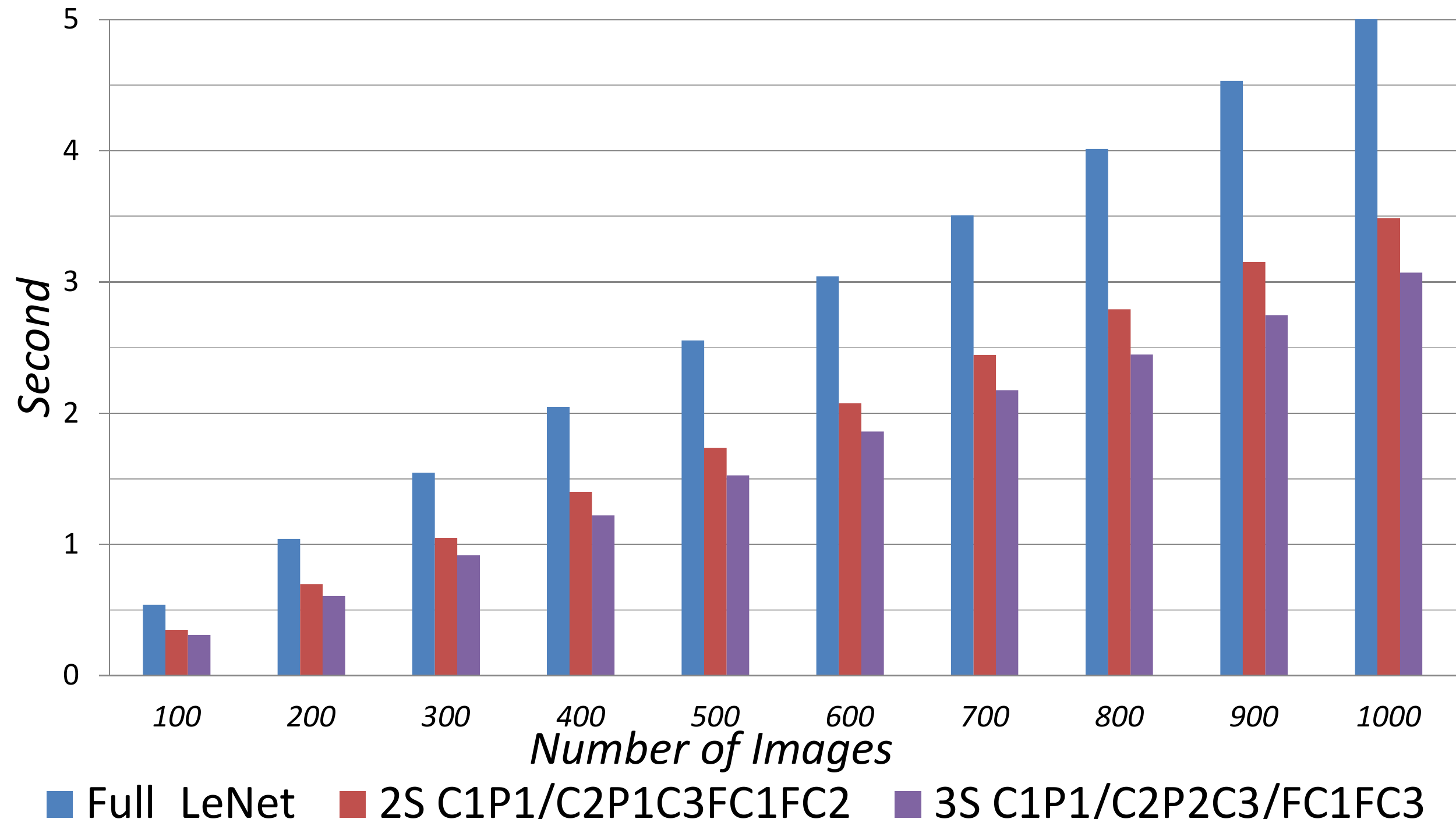}}
    \hfill
\subfloat[Comparison among worker node, two worker nodes, and three worker nodes for 1,000 to 10,000 images.
    \label{fig:3S10K}]
    {\includegraphics[width=0.49\linewidth]{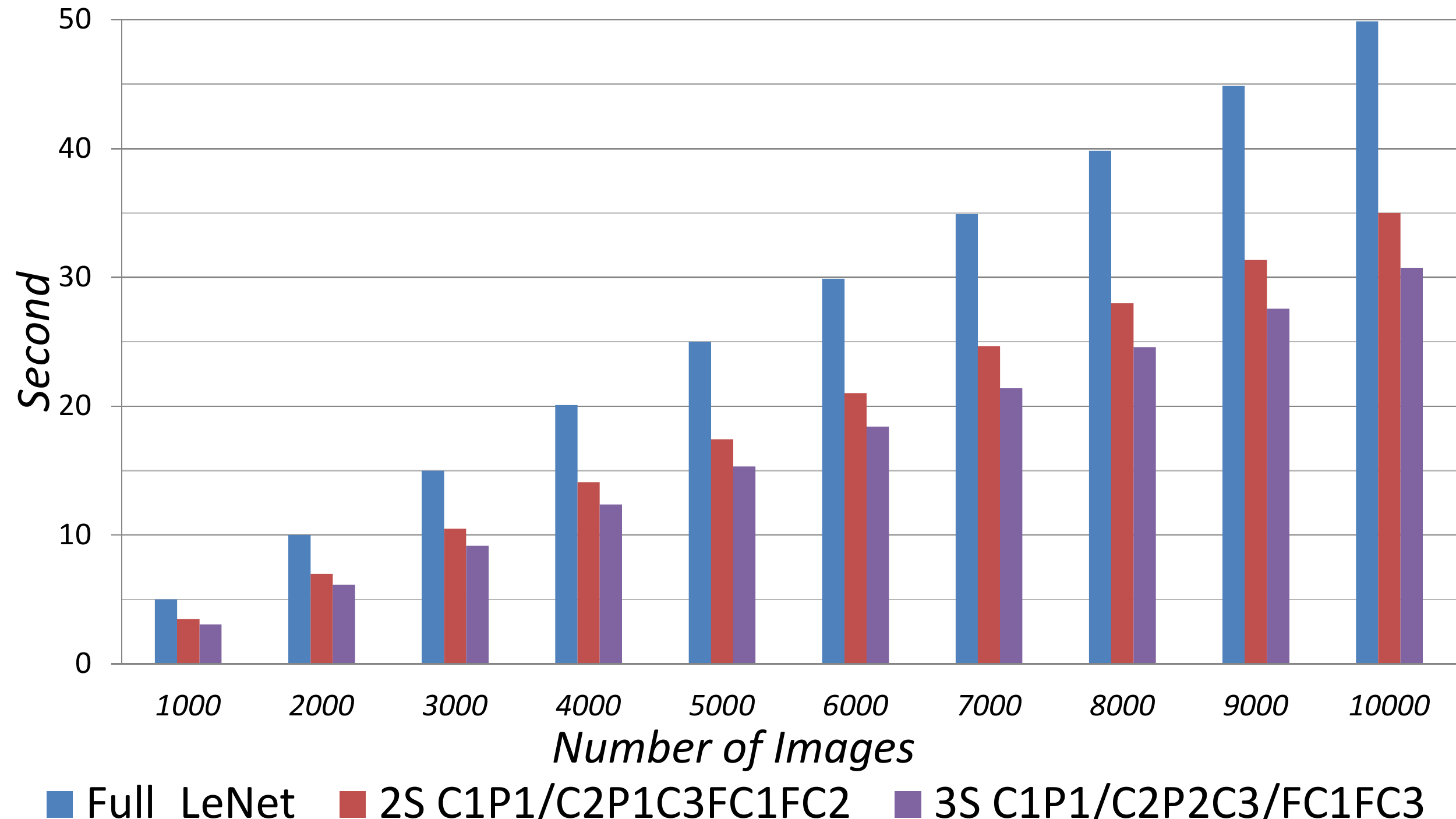}}
    \hfill
\vspace{3mm}
\caption{Comparison between pipeline and non-pipeline operation.}
\vspace{0mm}
\label{fig:2S3S_Result}
\end{figure}

\vspace{0mm}
\section{Comparison With State-of-the-Art}
\label{Comparison With State-of-the-Art}
\vspace{0mm}

Zhao \textit{et. al} \cite{zhao2018deepthings} proposes distributed intelligent video surveillance using distributed deep learning models.
This approach makes use of a multi-layer edge computing platform comprising of monitoring devices, cloud servers, and multilayer edge nodes.
This approach does not intimate how to partition the deep learning model based on the available edge nodes.
Li \textit{et. al} \cite{li2018edge} introduces a co-inference framework for CNN models to achieve low-latency edge intelligence.
This approach proposes the partitioning of CNN models on edge devices and the edge server to achieve inference.
This back and forth between the edge servers and edge devices introduces latency.

Mao \textit{et. al} \cite{mao2017modnn} proposes a local distributed computing approach that utilizes parallel computation for the convolution and fully connected layers of the CNN model.
Their approach uses a Biased One-Dimensional Partition scheme for the partitioning of convolution layers.
Mao et. al \cite{mao2017mednn} also achieves computation parallelism for the distributed edge computing system.
Their approach uses greedy two-dimensional partition to find the optimal partition strategy for splitting layers of the CNN models.
Both approaches use a specialized node to schedule and distribute workloads to other edge devices on the network.
After completion of operation at every node, the results are sent to the specialized node for re-scheduling and re-distribution.
In their approach, the back and forth between the specialized nodes and the other nodes on the network will increase the latency.
Operations in this technique are not parallel thereby leading to lower throughput.

\vspace{0mm}
\section{Conclusion}
\label{Conclusion}
\vspace{0mm}

In this paper, pipeline methodology has been utilized to share the computation and memory requirements for the inference of CNN.
This makes the CNN inference capable of deploying on resource-constrained devices without changing the CNN structure or sacrificing accuracy.
Also, it increases the throughput and makes it capable of use in real-time applications.
By sharing the CNN between two resource constraint devices, the inference can speed up by 1.55x, and by using three resource constraint devices the inference can speed up by 1.75x.
To the best of the authors' knowledge, this is the first work in deploying the CNN inference phase on multiple resource-constrained IoT node devices to achieve higher throughput.

\vspace{0mm}
\section{Acknowledgement }
\label{Acknowledgement }
\vspace{0mm}
Part of this work is supported by Faculty Research Grant 2020-2021 from the office of research at Tennessee Tech University.

\vspace{0mm}
\bibliographystyle{IEEEtran}
\bibliography{references}
\vspace{0mm}
\end{document}